\documentclass[10pt,twocolumn,letterpaper]{article}

\usepackage[pagenumbers]{cvpr} 
\usepackage{comment}
\usepackage{algorithm}
\usepackage{listings}
\usepackage{adjustbox}
\usepackage{tabularx}

\definecolor{cvprblue}{rgb}{0.21,0.49,0.74}
\usepackage[pagebackref,breaklinks,colorlinks,allcolors=cvprblue]{hyperref}

\def\paperID{} 
\def\confName{CVPR}
\def\confYear{2026}

\title{Multimodal Continual Instruction Tuning with Dynamic Gradient Guidance}

\author{
    Songze Li\textsuperscript{1}\quad
    Mingyu Gao\textsuperscript{1}\quad
    Tonghua Su\textsuperscript{1,2,3}\thanks{Corresponding author: thsu@hit.edu.cn. Code is available at \url{https://github.com/lisongze/DGG}.}\quad
    Xu-Yao Zhang\textsuperscript{4,5}\quad
    Zhongjie Wang\textsuperscript{1}
    \vspace{0.2cm}\\
    \textsuperscript{1}Harbin Institute of Technology \\ 
    \textsuperscript{2}Guangdong Laboratory of Artificial Intelligence and Digital Economy (SZ) \\
    \textsuperscript{3}Chongqing Research Institute of HIT \quad
    \textsuperscript{4} Institute of Automation, Chinese Academy of Sciences \\
    \textsuperscript{5}University of Chinese Academy of Sciences
}

\begin{document}
\maketitle
\begin{abstract}
Multimodal continual instruction tuning enables multimodal large language models to sequentially adapt to new tasks while building upon previously acquired knowledge. However, this continual learning paradigm faces the significant challenge of catastrophic forgetting, where learning new tasks leads to performance degradation on previous ones. 
In this paper, we introduce a novel insight into catastrophic forgetting by conceptualizing it as a problem of missing gradients from old tasks during new task learning. 
Our approach approximates these missing gradients by leveraging the geometric properties of the parameter space, specifically using the directional vector between current parameters and previously optimal parameters as gradient guidance. 
This approximated gradient can be further integrated with real gradients from a limited replay buffer and regulated by a Bernoulli sampling strategy that dynamically balances model stability and plasticity. 
Extensive experiments on multimodal continual instruction tuning datasets demonstrate that our method achieves state-of-the-art performance without model expansion, effectively mitigating catastrophic forgetting while maintaining a compact architecture. 
\end{abstract}
    
\section{Introduction}

\begin{figure}[t]
    \centering
    \includegraphics[width=\columnwidth]{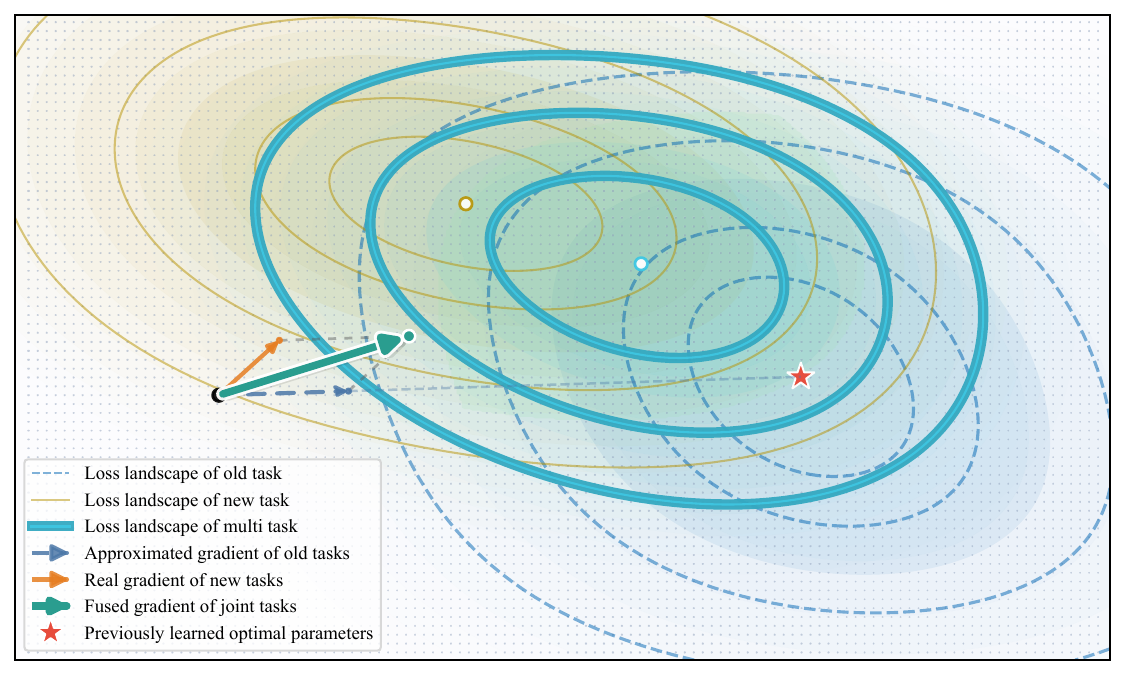}
    \caption{Illustration of our novel insight into catastrophic forgetting. We attribute catastrophic forgetting to the absence of old tasks' gradients during new task learning, which prevents gradient descent from converging to the optimal parameters achievable through joint training of all tasks. To address this problem, we approximate the missing gradients of old tasks by utilizing the optimal parameters from previous tasks (red star) as directional guides. The vector connecting current model parameters to these previously optimal parameters provides geometric guidance for approximating old task gradient directions. By integrating this approximated gradient with the new task gradient, we effectively simulate the joint training gradient, thereby alleviating catastrophic forgetting. To better convey our idea, all “gradients” in this and subsequent figure are plotted with the descent direction (the negative gradient) used by the optimizer.
}
    \label{fig:insight_of_forgetting}
    \vspace{-10pt}
\end{figure}	

In recent years, multimodal large language models (MLLMs) \cite{LLAVA, li2023blip, wu2024next} have garnered widespread attention for their remarkable ability to process and generate content across textual and visual modalities. 
These models typically follow a two-stage development paradigm: large-scale pre-training to establish cross-modal alignment through extensive datasets, followed by instruction tuning using carefully curated instruction-response pairs to enhance task-specific performance and instruction-following abilities.
By integrating vision and language processing capabilities, MLLMs have achieved impressive performance in diverse tasks such as image captioning \cite{chen2022visualgpt}, visual question answering \cite{LLAVA}, and multimodal reasoning \cite{zhang2023multimodal}. 

Despite their successes, the instruction tuning of MLLMs presents several challenges, particularly in the context of continual learning \cite{wang2024comprehensive, CLKD}. 
As these models are frequently finetuned with new instruction datasets, they risk forgetting previously acquired knowledge, a phenomenon known as catastrophic forgetting \cite{mccloskey1989catastrophic}. 
While retraining models from scratch with accumulated data can mitigate this issue, it becomes computationally prohibitive and environmentally unsustainable given the massive scale of modern MLLMs and the relentless influx of new data. 
These challenges have motivated the emerging field of multimodal continual instruction tuning (MCIT), which seeks to develop methods that enable MLLMs to acquire new skills continuously while preserving existing knowledge efficiently.

Many studies have explored MCIT, primarily building upon the LLaVA \cite{LLAVA} architecture with Low-Rank Adaptation (LoRA) \cite{LORA} for parameter-efficient fine-tuning. 
These methods typically leverage Mixture-of-Experts (MoE) \cite{MOE} structures and prompt tuning techniques to capture task-specific knowledge and maintain memory across different tasks \cite{CLMOE, HIDE, ModalPrompt}. 
However, such task-specific component learning inevitably leads to model expansion, introducing substantial additional parameter storage overhead and computational complexity during both the training and inference phases. 
Regularization-based approaches can be used to mitigate forgetting without model expansion by imposing constraints on parameter updates to preserve previously learned knowledge \cite{SEFE}. While effective to some extent, these methods typically rely on static regularization terms that remain fixed throughout the learning process, limiting their adaptability to the evolving optimization landscape.

In this paper, we propose an approach that enables learning new knowledge without model expansion while utilizing dynamic regularization to consolidate memory of previous tasks. 
We first revisit catastrophic forgetting and reformulate the challenge of knowledge preservation as a gradient approximation problem, offering a novel perspective to combat catastrophic forgetting.
We attribute the forgetting problem in continual learning to the absence of old tasks' gradients during optimization, which prevents gradient descent from converging to the optimal parameters achievable through joint learning of all tasks, consequently leading to performance degradation on previous tasks (see Fig. \ref{fig:insight_of_forgetting}). 
To approximate the missing gradients, we propose a dynamic gradient guidance method to approximate old tasks' gradients through directional guidance derived from the vector between current parameters and previously learned optimal parameters. 
This gradient guidance, which can be viewed as a regularization term, is dynamically adjusted throughout the learning process and intelligently combined with a limited replay buffer to provide a more accurate gradient approximation for old tasks.
Additionally, we introduce a Bernoulli sampling mechanism to dynamically regulate the application of these approximated gradients, enabling an effective balance between learning new tasks and preserving old knowledge. 
Our main contributions are as follows:
\begin{itemize}
    \item We provide new insights into catastrophic forgetting and reformulate knowledge preservation as an old tasks' gradients approximation problem.
    \item We propose a dynamic gradient guidance method to approximate old tasks' gradients, which can be combined with memory replay to achieve a more accurate gradient approximation.
    \item To balance model stability and plasticity, we develop a Bernoulli sampling-based dynamic gradient update strategy that dynamically controls the integration of approximated gradients.
    \item Experiments on two datasets demonstrate our method achieves state-of-the-art (SOTA) performance with a compact architecture, avoiding model expansion entirely.
\end{itemize}

\section{Related Work}
\textbf{Continual learning}, also known as lifelong learning or incremental learning, refers to the ability of machine learning models to acquire new knowledge from sequentially arriving data while retaining previously learned information.
Current continual learning methods can be broadly categorized into three main paradigms: replay-based, regularization-based and architecture-based \cite{parisi2019continual}.
Replay-based methods \cite{rebuffi2017icarl, castro2018end} maintain a subset of previous training samples, either in raw form or through generative models, and periodically revisit these samples during new task learning.
Regularization-based methods \cite{kirkpatrick2017overcoming, aljundi2018memory} alleviate forgetting by imposing constraints on parameter updates to protect important weights for previous tasks. 
Architecture-based methods \cite{aljundi2017expert} dynamically expand or modify the model structure to accommodate new knowledge while retaining previous knowledge.

\textbf{Multimodal continual instruction tuning} aims to endow MLLMs with the ability to learn from a stream of instruction-following tasks without forgetting previously acquired knowledge. 
Recently, this challenging problem has attracted significant research interest, with most approaches building upon MoE architectures to preserve sequential knowledge, albeit at the cost of model expansion.
For example, CoIN \cite{COIN} proposes MoELoRA to acquire distinct knowledge for different tasks.
CL-MoE \cite{CLMOE} introduces a Dual-Router MoE for precise expert activation and a Momentum MoE for dynamic expert updating.
HiDE \cite{HIDE} employs  a task-specific expansion and task-general fusion framework, which decouples the learning process hierarchically.
DISCO \cite{DISCO} proposes a dynamic knowledge organization and subspace selective activation framework to address challenges in federated continual instruction tuning scenarios. 
Beyond the MoE paradigm, ModalPrompt \cite{ModalPrompt} reduces forgetting and computational complexity through efficient prompt fusion, but still suffers from model expansion.
SEFE \cite{SEFE} introduces RegLoRA, which addresses essential forgetting by imposing regularization constraints on critical elements within the weight update matrices.

\begin{figure*}[t]
    \centering
    \begin{subfigure}{0.33\textwidth}  
        \centering
        \includegraphics[width=\textwidth]{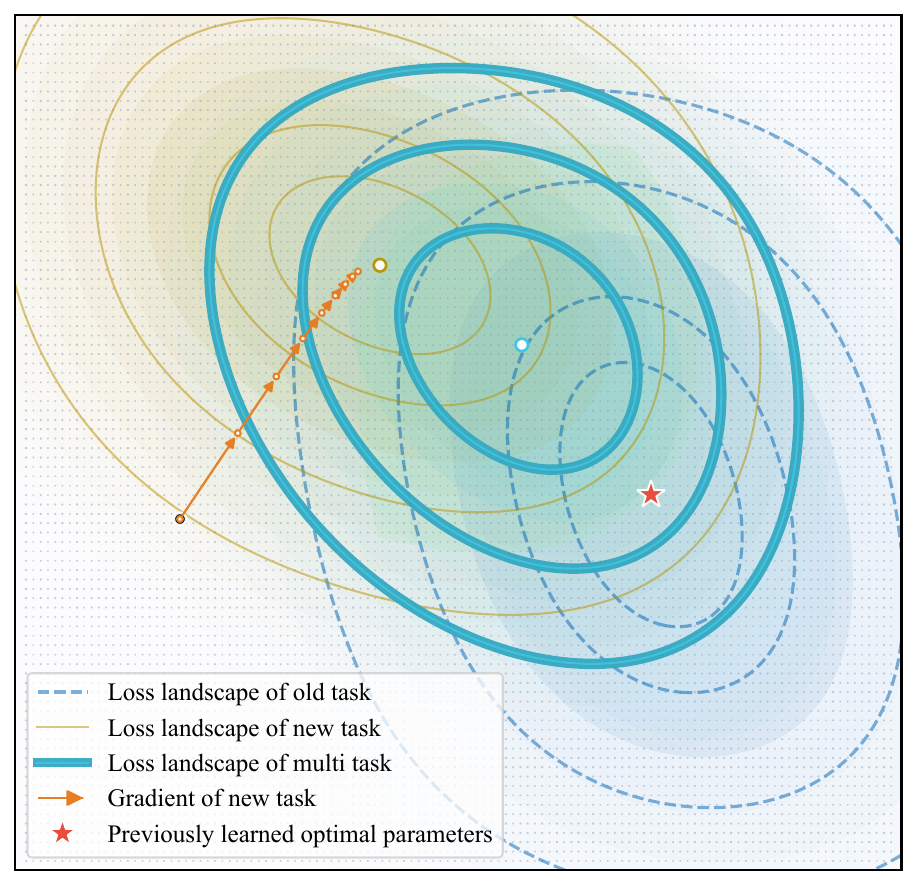}  
        \caption{Learning without memory retention}
        \label{fig:optimization_with_nothing}
    \end{subfigure}
    \begin{subfigure}{0.33\textwidth}  
        \centering
        \includegraphics[width=\textwidth]{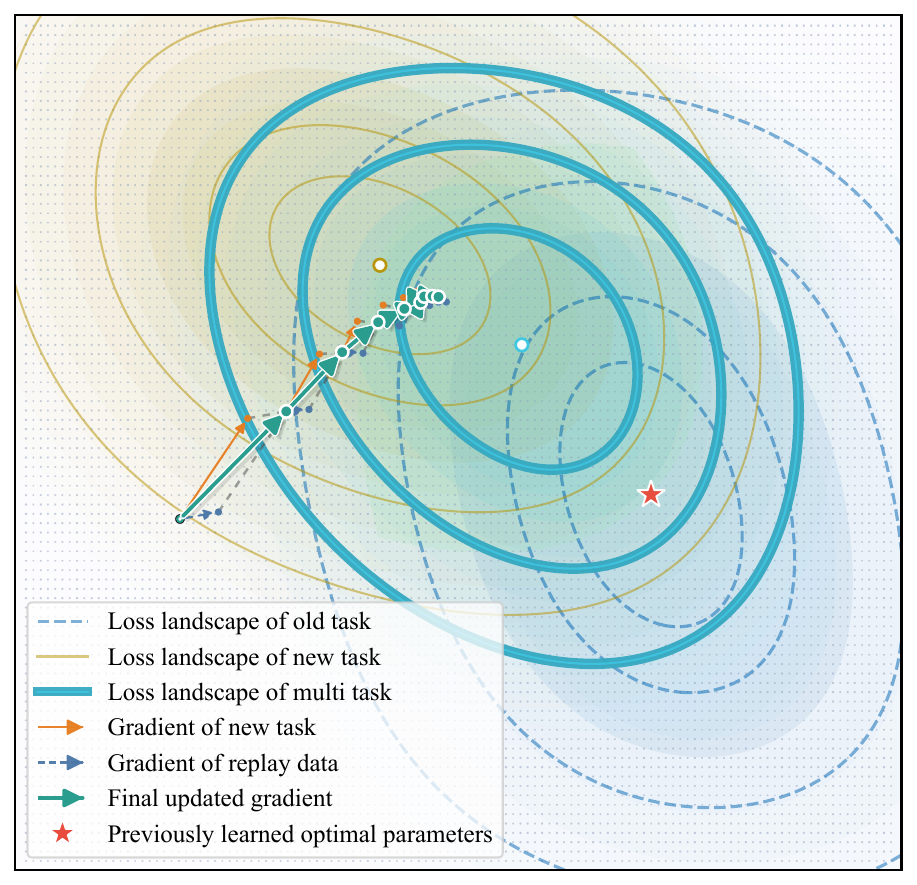}  
        \caption{Learning with replay data}
        \label{fig:optimization_with_replay}
    \end{subfigure}
    \begin{subfigure}{0.33\textwidth}  
        \centering
        \includegraphics[width=\textwidth]{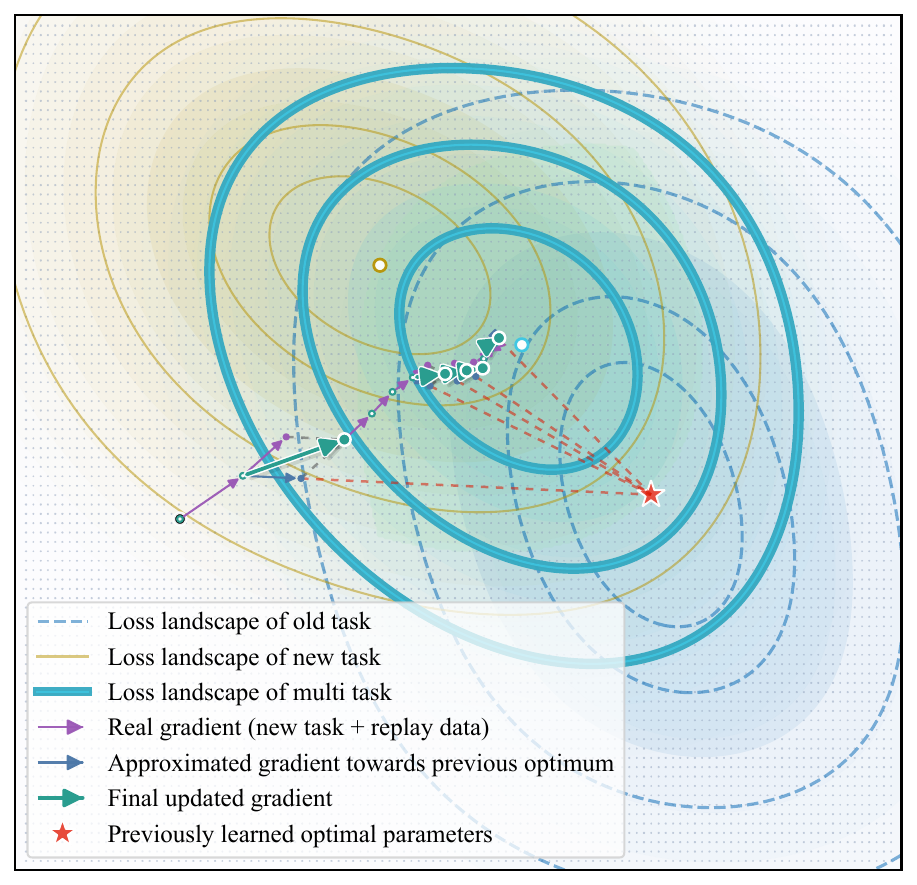}  
        \caption{Learning with our method}
        \label{fig:optimization_with_gradient_app}
    \end{subfigure}
    
    \caption{Optimization process with different memory retention strategy. (a) Learning a new task without any memory retention tricks. Due to the exclusive presence of new task gradients (yellow arrow) and the absence of old task gradients, the model converges directly to the optimal solution for the new task, resulting in complete forgetting of previous knowledge. (b) Learning a new task with replay data. The inclusion of a limited number of replay samples provides partial gradient information from old tasks (blue arrow), enabling the model to converge to parameters that retain some memory. However, the gradients from these samples cannot represent the expected gradient over the entire old task dataset throughout the optimization process, leading to suboptimal convergence relative to multi-task learning and residual catastrophic forgetting. (c) Learning a new task with our dynamical gradient guidance. Our method approximates old task gradients by leveraging optimal parameters from previous tasks as directional guides (blue arrow), fused with real gradients from cached replay samples (purple arrow, combined with new task gradient). This approximation is dynamically regulated through Bernoulli sampling (red dotted line) to control gradient update frequency, achieving balanced convergence towards joint task optimization.}
    \label{fig:optimization_compare}
\end{figure*}

\section{Preliminary}

The problem of MCIT focuses on adapting MLLMs to evolving tasks while preserving previously learned capabilities. 
In this setting, a model parameterized by $\theta$ encounters a sequence of distinct tasks $\{\mathcal{T}_1, \mathcal{T}_2, ..., \mathcal{T}_T\}$ in chronological order, building upon prior knowledge from multimodal pre-training.
Each task $\mathcal{T}_t$ consists of a collection of multimodal examples:
\begin{equation}
\mathcal{T}_t = \{x_k^{(i)}=(v_t^{(i)}, q_t^{(i)}, a_t^{(i)})\}_{i=1}^{\left|\mathcal{T}_t\right|}, \quad \forall t \in \{1,...,T\},
\end{equation}
where $v_t^{(i)}$ represents visual inputs, $q_t^{(i)}$ denotes instructional queries, and $a_t^{(i)}$ corresponds to target responses for the $i$-th instance in task $t$, with $\left|\mathcal{T}_t\right|$ indicating the task's dataset size.

The learning objective for each task follows an autoregressive formulation. For a given input sequence, the model optimizes:
\begin{equation}
\mathcal{L}(\theta; \mathcal{T}_t) = \mathbb{E}_{(v,q,a) \sim \mathcal{T}_t}\left[\sum_{j=1}^{|a|} -\log P(a_j | v, q, a_{<j}; \theta)\right],
\end{equation}
where $|a|$ denotes the length of the target response.

Under the continual learning paradigm, when training on task $\mathcal{T}_t$, the ideal objective is to minimize the composite loss over all encountered tasks. Since the transformer architecture employs LayerNorm rather than BatchNorm, and the loss function contains no additional components such as contrastive learning objectives, the overall loss can be decomposed as a simple summation over individual samples. Specifically, the loss function for task $k$ can be expressed as:

\begin{equation}
\mathcal{L}(\theta; \mathcal{T}_k) = \frac{1}{\left|\mathcal{T}_k\right|} \sum_{i=1}^{\left|\mathcal{T}_k\right|} \ell\left(x_k^{(i)}; \theta\right),
\end{equation}
where $\ell\left(x_k^{(i)}; \theta\right)$ denotes the negative log-likelihood loss for the $i$-th sample of task $k$. Consequently, the composite loss across all tasks from $1$ to $t$ becomes:

\begin{equation}
\begin{aligned}
\mathcal{L}(\theta; \sum_{k=1}^{t} \cup \mathcal{T}_k)) &= \frac{1}{\left|\mathcal{T}_{1:t}\right|} \sum_{k=1}^{t} \sum_{i=1}^{\left|\mathcal{T}_k\right|} \ell\left(x_k^{(i)}; \theta\right) \\ &= \sum_{k=1}^{t} \lambda_k \mathcal{L}(\theta; \mathcal{T}_k),
\end{aligned}
\end{equation}
where $\lambda_k = \frac{\left|\mathcal{T}_k\right|}{\left|\mathcal{T}_{1:t}\right|}$ represents the relative sample size of task $k$ $\left|\mathcal{T}_{k}\right|$ compared to $\left|\mathcal{T}_{1:t}\right|$ which denotes the total samples from all tasks up to $t$. 
Assuming each task contains an equal number of samples, $\lambda_k$ becomes a constant value that can be omitted from the formulation. 
Hence we have:
\begin{equation}
\label{eq:loss_joint}
\mathcal{L}(\theta; \sum_{k=1}^{t} \cup \mathcal{T}_k)) = \sum_{k=1}^{t} \mathcal{L}_k(\theta; \mathcal{T}_k).
\end{equation}

The primary challenge in continual learning stems from the unavailability of previous tasks' data $\{\mathcal{T}_1, \ldots, \mathcal{T}_{t-1}\}$ during training on $\mathcal{T}_t$, which leads to catastrophic forgetting. 
A common approach to address this issue involves maintaining a replay buffer $\mathcal{M}$ containing representative samples from previous tasks, which are periodically revisited during training to reinforce the model's memory of earlier acquired knowledge.

\section{Method}
\subsection{New Insight into Catastrophic Forgetting}

We attribute the catastrophic forgetting problem to the absence of old task gradients during optimization, which prevents gradient descent from converging to the optimal parameters achievable through joint learning of all tasks.
Consider a model parameterized by $\theta$, trained on two datasets $\mathcal{T}_1$ and $\mathcal{T}_2$. According to Eq. \ref{eq:loss_joint}, we have the loss function for the joint dataset $\mathcal{T}_1 \cup \mathcal{T}_2$ as:
\begin{equation}
   \mathcal{L}(\theta; \mathcal{T}_1 \cup \mathcal{T}_2) = \mathcal{L}(\theta; \mathcal{T}_1) +\mathcal{L}(\theta; \mathcal{T}_2), 
\end{equation}

By the linearity of differentiation, the gradient with respect to the model parameters $\theta$ also satisfies:
\begin{equation}
\nabla_\theta \mathcal{L}(\theta; \mathcal{T}_1 \cup \mathcal{T}_2)
= \nabla_\theta \mathcal{L}(\theta; \mathcal{T}_1)
+ \nabla_\theta \mathcal{L}(\theta; \mathcal{T}_2).
\end{equation}

Suppose we are currently engaged in continual learning and have progressed to the second task $\mathcal{T}_2$, where we can compute both the loss $\mathcal{L}(\theta; \mathcal{T}_2)$ and its corresponding gradient $\nabla_\theta \mathcal{L}(\theta; \mathcal{T}_2)$ (see Fig. \ref{fig:optimization_with_nothing}). 

However, to maintain memory of the old task $\mathcal{T}_1$ during continual learning, we need to learn the parameters that achieve: 
\begin{equation}
    \theta^*_{1:2} = \mathop{\arg\min}\limits_{\theta} \mathcal{L}(\theta; \mathcal{T}_1 \cup \mathcal{T}_2) .
\end{equation}

Unfortunately, since the data from the previous task $\mathcal{T}_1$ is no longer accessible, we cannot directly compute $\mathcal{L}(\theta; \mathcal{T}_1)$ or its gradient $\nabla_\theta \mathcal{L}(\theta; \mathcal{T}_1)$. 
The core idea of our approach is that if we can accurately approximate $\nabla_\theta \mathcal{L}(\theta; \mathcal{T}_1)$, we would be able to obtain $\nabla_\theta \mathcal{L}(\theta; \mathcal{T}_1 \cup \mathcal{T}_2)$ and subsequently employ gradient-based optimization to find $\theta^*_{1:2}$, thereby effectively preserving the memory of $\mathcal{T}_1$.

\begin{algorithm}[t]
\caption{Pseudocode of our method in a PyTorch-like style.}
\label{alg:code}
\definecolor{codeblue}{rgb}{0.25,0.5,0.5}
\lstset{
  backgroundcolor=\color{white},
  basicstyle=\fontsize{7.2pt}{7.2pt}\ttfamily\selectfont,
  columns=fullflexible,
  breaklines=true,
  captionpos=b,
  commentstyle=\fontsize{7.2pt}{7.2pt}\color{codeblue},
  keywordstyle=\fontsize{7.2pt}{7.2pt},
}
\begin{lstlisting}[language=python]
# model: Current model with parameters to be updated
# optimal_params: Dictionary of optimal parameters from previously learned tasks
# alpha: Bernoulli sampling probability for gradient update
    
# Sample from Bernoulli distribution to decide whether to update
if Bernoulli(alpha) == 1:
    
    # Iterate through all model parameters
    for param_name, current_param in model.named_parameters():
        
        # Skip parameters without gradients
        if current_param.grad is None:
            continue
            
        # Get current gradient and compute its norm
        current_grad = current_param.grad
        current_grad_norm = current_grad.norm()
        
        # Skip if gradient norm is zero
        if current_grad_norm == 0:
            continue
            
        # Compute directional vector between current and optimal parameters
        optimal_param = optimal_params[param_name]
        direction_vector = current_param - optimal_param
        direction_norm = direction_vector.norm()
        
        # Scale directional vector if its norm exceeds gradient norm
        if direction_norm >= current_grad_norm:
            # Normalize and scale to match gradient magnitude
            scaled_direction = (direction_vector / direction_norm) * current_grad_norm
            current_param.grad += scaled_direction
        else:
            # Use original directional vector
            current_param.grad += direction_vector
\end{lstlisting}
\end{algorithm}

\subsection{Approximation with Gradient Guidance}
The core idea to mitigate forgetting is to accurately approximate $\nabla_\theta \mathcal{L}(\theta; \mathcal{T}_1)$.
A straightforward and commonly adopted approach is to cache a subset of samples $\mathcal{M}$ from the old task and replay them to estimate the gradient for $\mathcal{T}_1$. 
However, due to the limited number of cached samples from the old task, it is generally difficult to accurately approximate the entire data distribution of $\mathcal{T}_1$. 
Consequently, the gradient estimated via replaying the cached samples $\mathcal{M}$ may not represent the expected gradient over the entire dataset of $\mathcal{T}_1$ throughout the gradient descent learning process. 
As a result, continual learning methods relying on replay often introduce bias, favoring the current task and exhibiting limited ability to retain memory of old tasks (see Fig. \ref{fig:optimization_with_replay}). 
Therefore, it is necessary to seek an approximation that more closely resembles the expected gradient of $\mathcal{T}_1$ over the entire gradient descent process, and utilize it to approximate the current gradient for $\mathcal{T}_1$.

Our approach leverages the optimal parameters obtained from training on previous tasks as a guidance to compute an approximate gradient for old tasks.
For example, in the process of learning the first task, we optimize $\theta$ through gradient descent algorithms to obtain an optimal parameter set for $\mathcal{T}_1$, specifically:
\begin{equation}
\theta^*_1 = \mathop{\arg\min}\limits_{\theta} \mathcal{L}(\theta; \mathcal{T}_1).
\end{equation}

\begin{table*}[t]
\centering
\caption{Experimental results on VQAv2 dataset with 0.5k replay samples per task.}
\label{tab:VQAv2_main}
\begin{tabularx}{0.98\textwidth}{l|*{10}{>{\centering\arraybackslash}X}|c}
\toprule
Method      & Rec.  & Loc.  & Jud.  & Com.  & Cou.  & Act.  & Col.  & Typ.  & Sub.  & Cau.      & FAA  \\ \hline
MultiTask & 55.15 & 41.88 & 80.74 & 75.47 & 49.81 & 75.97 & 73.03 & 61.02 & 60.54 & 29.49 &  66.26 \\ \hline\hline
Ours        & \textbf{55.55} & \textbf{41.03} & \textbf{78.67} & \textbf{76.12} & 48.33 & \textbf{75.62} & 69.20  & \textbf{61.19} & \textbf{60.35} & 28.11 &  \textbf{65.17} \\ \hline\hline
CL-MoE      & 46.50 & 37.18 & 75.22 & 71.39 & 40.90  & 69.54 & 43.66 & 52.68 & 55.55 & 20.74 &  57.27 \\
HiDE        & 49.27 & 33.62 & 72.27 & 69.11 & 43.72 & 70.17 & 65.36 & 55.24 & 56.42 & 25.81 &  59.44 \\
SEFE        & 50.55 & 39.46 & 78.42 & 75.96 & 48.43 & 72.86 & 70.50  & 58.05 & 58.54 & 29.95 &  63.57 \\
DISCO       & 54.48 & 38.60 & 73.98 & 68.22 & \textbf{49.23} & 72.65 & \textbf{72.91} & 59.62 & 57.61 & \textbf{29.95} &  63.09 \\ 
\bottomrule
\end{tabularx}
\end{table*}

Throughout the gradient descent optimization process, the algorithm ultimately converges toward the target $\theta^*_1$. Consequently, the direction pointing toward $\theta^*_1$ can, to some extent, reflect the expected gradient direction throughout the optimization trajectory. Building upon this intuition, we propose an approximation method $\hat{g}$. Assuming $\theta$ represents the current model parameters during the learning of $\mathcal{T}_2$, we approximate the gradient for $\mathcal{T}_1$ using $\theta - \theta^*_1$ (see Fig. \ref{fig:optimization_with_gradient_app}). However, since $\theta - \theta^*_1$ only indicates the gradient direction and its direct application might lead to excessively large gradient magnitudes, we need to scale it appropriately. This scaling can be achieved by utilizing the magnitude of the $\mathcal{T}_2$ gradient:
\begin{equation}
\label{eq:g_hat}
\hat{g} = 
\begin{cases} 
\frac{\theta - \theta^*_1}{\|\theta - \theta^*_1\|} \cdot \|\nabla_\theta \mathcal{L}(\theta; \mathcal{T}_2)\|, & 
    \begin{aligned}
    &\text{if } \|\theta - \theta^*_1\| > \\
    &\|\nabla_\theta \mathcal{L}(\theta; \mathcal{T}_2)\|
    \end{aligned} \\
\theta - \theta^*_1, & \text{otherwise}
\end{cases}.
\end{equation}

Furthermore, we can also introduce the replay data $\mathcal{M}$ to compute the real gradients for $\mathcal{T}_1$, thereby achieving a more accurate approximation of $\nabla_\theta \mathcal{L}(\theta; \mathcal{T}_1)$, namely:
\begin{equation}
    \nabla_\theta \mathcal{L}(\theta; \mathcal{T}_1) \approx \hat{g} + \nabla_\theta \mathcal{L}(\theta; \mathcal{M}).
\end{equation}

Finally, we update the model with following gradient:
\begin{equation}
\begin{aligned}
    \nabla_\theta \mathcal{L}(\theta; \mathcal{T}_1 \cup \mathcal{T}_2) 
    & \approx \hat{g} + \nabla_\theta \mathcal{L}(\theta; \mathcal{M}) + \nabla_\theta \mathcal{L}(\theta; \mathcal{T}_2) \\
    & \approx \hat{g} + \nabla_\theta \mathcal{L}(\theta; \mathcal{T}_2 \cup \mathcal{M}).
\end{aligned}
\end{equation}

For subsequent tasks $t \left(t \geq 2\right)$, we can treat all old tasks as a joint task. Therefore, we have:
\begin{equation}
    \nabla_\theta \mathcal{L}(\theta; \sum_{i=1}^{t}\cup \mathcal{T}_i) = \nabla_\theta \mathcal{L}(\theta; \sum_{i=1}^{t-1}\cup \mathcal{T}_i) + \nabla_\theta \mathcal{L}(\theta; \mathcal{T}_t).
\end{equation}
Then we can compute $\hat{g}$ by leveraging the continually learned optimal parameters $\theta^*_{1:t-1}$ from previous tasks and the current task gradient $\nabla_\theta \mathcal{L}(\theta; \mathcal{T}_t)$, and update the model as follow:
\begin{equation}
\label{eq:final_gradient_update}
    \nabla_\theta \mathcal{L}(\theta; \sum_{i=1}^{t}\cup \mathcal{T}_i) 
    \approx \hat{g} + \nabla_\theta \mathcal{L}(\theta; \mathcal{T}_t \cup \mathcal{M}),
\end{equation}
where
\begin{equation}
\label{eq:g_hat_final}
\hat{g} = 
\begin{cases} 
\frac{\theta - \theta^*_{1:t-1}}{\|\theta - \theta^*_{1:t-1}\|} \cdot \|\nabla_\theta \mathcal{L}(\theta; \mathcal{T}_t)\|, & 
    \begin{aligned}
    &\text{if } \|\theta - \theta^*_{1:t-1}\| > \\
    &\|\nabla_\theta \mathcal{L}(\theta; \mathcal{T}_t)\|
    \end{aligned} \\
\theta - \theta^*_{1:t-1}, & \text{otherwise}
\end{cases}.
\end{equation}

\subsection{Dynamic Gradient Update with Bernoulli Sampling}
Now we can naturally integrate Eq. \ref{eq:final_gradient_update} with gradient descent (GD) \cite{robbins1951stochastic} for parameter optimization. However, when employing GD for optimization, the random sampling of mini-batches introduces inherent stochasticity in gradient updates. 
Furthermore, excessive updates to the gradients of old tasks may cause the model to become overly biased towards previous tasks, reducing its plasticity and thereby impairing its ability to learn new tasks.
To emulate the stochastic nature of gradient descent and regulate the update frequency of old task gradients, thereby preventing the model from overfitting to previous knowledge and facilitating effective learning of new tasks, we introduce a Bernoulli sampling-based dynamic gradient update mechanism. 
The Bernoulli distribution is a discrete probability distribution characterized by a single probability parameter $\alpha$, which represents the probability of a binary outcome (success or failure). In our method, we define a Bernoulli random variable with parameter $\alpha$ to stochastically determine whether to incorporate the approximated old task gradient $\hat{g}$ during optimization. 

Specifically, at each optimization step, we sample from this distribution. If the outcome is 1, we update the model parameters using both the approximated old task gradient $\hat{g}$ and the gradient from the current task and replay data; otherwise, we update using only the latter. This dynamic update rule is formally defined as:
\begin{equation}
\nabla_\theta \mathcal{L}\left(\theta; \cup_{i=1}^t \mathcal{T}_i\right) =
\begin{cases} 
\hat{g} + \nabla_\theta \mathcal{L}(\theta; \mathcal{T}_t \cup \mathcal{M}), & \text{if } \mathcal{B}(\alpha) = 1 \\
\nabla_\theta \mathcal{L}(\theta; \mathcal{T}_t \cup \mathcal{M}), & \text{if } \mathcal{B}(\alpha) = 0
\end{cases},
\end{equation}
where $\mathcal{B}(\alpha)$ denotes the Bernoulli random variable and $\alpha$ represents the success probability.

By controlling the frequency of old task gradient updates , our method effectively balances model plasticity (adaptation to new tasks) and stability (retention of old task knowledge), mitigating catastrophic forgetting while maintaining learning efficiency. Algorithm \ref{alg:code} provides the pseudo-code of our method.

\begin{table*}[t]
\centering
\caption{Experimental results on UCIT dataset with 2k replay samples per task.}
\label{tab:UCIT_main}

\begin{tabularx}{0.92\textwidth}{l|*{6}{>{\centering\arraybackslash}X}|c}
\toprule
Method      & ImageNet-R        & ArxivQA        & VizWiz         & IconQA         & CLEVR          & Flickr30k      & FAA            \\ \hline
MultiTask & 90.63          & 91.30          & 61.81          & 73.90          & 73.60          & 57.45          & 74.78          \\ \hline\hline
Ours        & \textbf{91.07} & 91.37 & \textbf{59.40} & \textbf{73.03} & \textbf{71.67} & 56.35          & \textbf{73.82} \\ \hline\hline
CL-MoE      & 66.33          & 77.00          & 44.78          & 51.87          & 53.53          & 57.42          & 58.49          \\
HiDE        & 84.03          & 90.73          & 44.43          & 58.93          & 41.37          & 54.25          & 62.29          \\
SEFE        & 80.83          & 78.00          & 47.01          & 69.63          & 65.83          & \textbf{57.92} & 66.54          \\
DISCO       & 87.43          & \textbf{93.07}          & 46.96          & 68.13          & 65.70          & 56.69          & 69.66          \\ \bottomrule
\end{tabularx}
\end{table*}

\section{Experiments}

\subsection{Experimental Setup}
\textbf{Datasets and Baselines.} 
We conducted training and evaluation on two MCIT datasets. First, we utilized VQAv2 \cite{VQAv2}, following the setup of CL-MoE \cite{CLMOE}, which partitions the dataset into 10 subtasks based on question types: Recognition, Location, Judge, Commonsense, Count, Action, Color, Type, Subcategory, and Causal. 
The second dataset is the more challenging UCIT dataset \cite{HIDE}, which comprises 6 distinct datasets with significant differences in image data distributions: ImageNet-R \cite{ImageNetR}, ArxivQA \cite{ArxivQA}, VizWiz-caption \cite{VizWiz}, IconQA \cite{IconQA}, CLEVR-Math \cite{CLEVR}, and Flickr30k \cite{Flickr30k}.
For both datasets, we compared our method against several recent SOTA MCIT approaches, including CL-MoE \cite{CLMOE}, SEFE \cite{SEFE}, HiDE \cite{HIDE}, and DISCO \cite{DISCO}. 

\noindent\textbf{Evaluation Metrics.}
Regarding evaluation metrics, we followed HiDE in reporting the final average accuracy (FAA) across all learned tasks after completing the final task. However, since the test sample sizes vary across different tasks in VQAv2, directly averaging per-task accuracy would be unfair. 
Therefore, we report the FAA based on the actual number of test samples per task, calculated as:
\begin{equation}
    \text{FAA}=\sum_{i=1}^T\frac{\left| \mathcal{T}_i\right|}{\left| \mathcal{T}_{1:T}\right|}a^T_i,
\end{equation}
where $a^T_i$ indicates the accuracy of the $i$-th task after completing the learning of the final task $T$.

\noindent\textbf{Implementation Details.}
All experiments are built upon the LLaVA-7B MLLM and employ LoRA for instruction tuning. 
For the VQAv2 dataset, we set the LoRA rank to 128, while for the UCIT dataset we use a rank of 48. The continual instruction tuning task sequence for VQAv2 follows the order: Recognition $\rightarrow$ Location $\rightarrow$ Judge $\rightarrow$ Commonsense $\rightarrow$ Count $\rightarrow$ Action $\rightarrow$ Color $\rightarrow$ Type $\rightarrow$ Subcategory $\rightarrow$ Causal. During continual instruction tuning, each task caches 500 (0.5k) samples for replay, and all tasks share a consistent Bernoulli probability $\alpha$ of 0.2. For the UCIT dataset, the task sequence is: ImageNet-R $\rightarrow$ ArxivQA $\rightarrow$ VizWiz $\rightarrow$ IconQA $\rightarrow$ CLEVR $\rightarrow$ Flickr30k, with each task caching 2,000 (2k) samples for replay. The task-specific Bernoulli parameters are configured as follows: ArxivQA (0.1), VizWiz (0.1), IconQA (0.05), CLEVR (0.05), and Flickr30k (0.1). More details are presented in appendix.

\begin{figure}[t]
    \centering
    \begin{subfigure}{0.23\textwidth}  
        \centering
        \includegraphics[width=\textwidth]{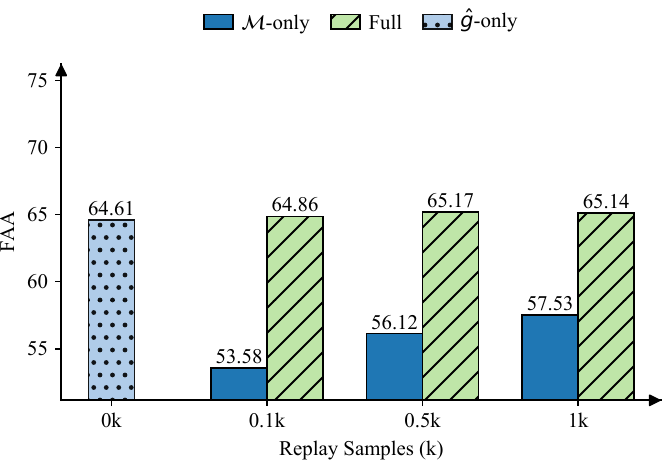}  
        \caption{VQAv2}
        \label{fig:ablation_grad_app_vqav2}
    \end{subfigure}
    \begin{subfigure}{0.23\textwidth}  
        \centering
        \includegraphics[width=\textwidth]{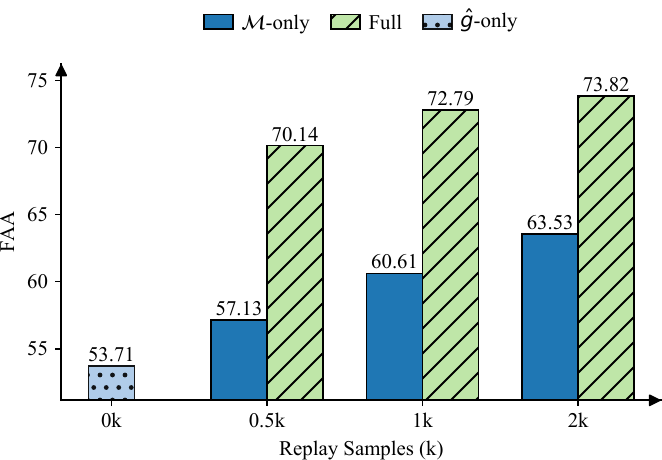}  
        \caption{UCIT}
        \label{fig:ablation_grad_app_ucit}
    \end{subfigure}
    
    \caption{Result of ablation on gradient approximation. We conduct this ablation under two configurations: using only replay buffers without $\hat{g}$ ($\mathcal{M}$-only) and using only $\hat{g}$ without replay buffers ($\hat{g}$-only). Full represents the full version of our method which integrates both of $\hat{g}$ and replay buffer $\mathcal{M}$.}
    \label{fig:ablation_grad_app}
\end{figure}

\subsection{Main Results}
The experimental results on the VQAv2 and UCIT datasets are summarized in Tables~\ref{tab:VQAv2_main} and~\ref{tab:UCIT_main}, respectively. 
All baseline methods are evaluated using the MCITlib benchmarking framework \cite{guo2025mcitlib}, with MultiTask learning serving as the performance upper bound. 
Our method achieves SOTA performance on both datasets among all baselines. On VQAv2, it attains 65.17\% FAA, outperforming the strongest baseline SEFE (63.57\% FAA) by 1.60\%. 
Notably, our method demonstrates superior performance on specific tasks including Recognition (55.55\%), Commonsense (76.12\%), and Type (61.19\%), even surpassing the MultiTask upper bound in certain categories while slightly underperforming on Color (69.20\%) and Causal (28.11\%) tasks compared to some baselines. On the more challenging UCIT dataset, which comprises 6 tasks with significant distribution shifts, our method achieves 73.82\% FAA, exceeding the strongest baseline DISCO (69.66\% FAA) by 4.16\%. 

Remarkably, our method demonstrates highly competitive performance compared to the MultiTask upper bound, with minimal gaps of 1.09\% in FAA on VQAv2, and 0.96\% in FAA on UCIT.
This achievement is particularly significant considering that most of compared baselines employ MoE architectures to learn task-specific parameters, whereas our approach directly addresses the continual instruction tuning at the optimization level without requiring specialized model components. By effectively approximating gradients for previous tasks within the same parameter space, our method provides a more elegant and efficient solution for knowledge retention.

\begin{table}[]
\caption{Ablation on task sequence (VQAv2 $\rightarrow$ VizWiz
$\rightarrow$ TextVQA $\rightarrow$ Flickr30k). Each task caches 0.5k samples for replay. All tasks share a consistent Bernoulli probability $\alpha$ of 0.1.}
\label{tab:ablation_gradient_app_new_sequence}
\resizebox{0.48\textwidth}{!}{
\begin{tabular}{l|cccc|cc}
\toprule
                   & VQAv2 & VizWiz & TextVQA & Flickr30k & FAA  \\ \hline
MultiTask          & 67.48 & 62.47  & 54.10   & 57.07     & 66.95 \\ \hline\hline 
Full               & 65.12 & 57.84  & 51.70   & 54.57     & 64.55 \\  
$\hat{g}$-only     & 62.54 & 54.94  & 52.50   & 53.92     & 62.06 \\  
$\mathcal{M}$-only & 58.16 & 53.90   & 43.46   & 57.91      & 57.75 \\ \bottomrule
\end{tabular}
}
\end{table}

\subsection{Ablation}
\textbf{Ablation on Gradient Approximation.} To evaluate the individual contributions of our two gradient approximation strategies -- the gradient guidance approximation $\hat{g}$ computed from optimal old task parameters and the real gradient computed from cached samples $\mathcal{M}$ -- we conduct ablation studies under two configurations: using only replay buffers without $\hat{g}$ ($\mathcal{M}$-only) and using only $\hat{g}$ without replay buffers ($\hat{g}$-only). We further investigate three different buffer sizes for each task: 0.1k, 0.5k, and 1k for VQAv2; 0.5k, 1k, and 2k for UCIT.
As shown in Fig. \ref{fig:ablation_grad_app}, the results reveal distinct patterns across datasets. On VQAv2, $\hat{g}$ plays a dominant role in memory preservation, achieving 64.61\% FAA even without any replay data, which surpasses the best baseline performance. In contrast, relying solely on replay buffers with 1k samples yields only 57.73\% FAA, significantly lower than using $\hat{g}$ alone. 
Conversely, on UCIT, replay buffers demonstrate greater importance for knowledge retention. 
Even with only 0.5k samples, $\mathcal{M}$-only achieve 57.13\% FAA, outperforming the $\hat{g}$-only approach (53.71\% FAA).
We hypothesize that this discrepancy stems from differences in data distribution characteristics. 
While VQAv2 contains tasks from the same visual domain, UCIT comprises 6 distinct datasets with substantial distribution shifts.

To validate this hypothesis, we extract two tasks from UCIT, VizWiz and Flickr30k, which exhibit similar data distributions, and incorporate two additional datasets, VQAv2 and TextVQA \cite{TextVQA}, that share analogous visual characteristics (see more in appendix), thereby forming a new task sequence: VQAv2 $\rightarrow$ VizWiz $\rightarrow$ TextVQA $\rightarrow$ Flickr30k. 
As shown in Table \ref{tab:ablation_gradient_app_new_sequence},  the $\hat{g}$-only approach significantly outperforms $\mathcal{M}$-only methods by  4.31\%. 
These results confirm that large distribution shifts impair the approximation accuracy of $\hat{g}$, leading to increased reliance on the replay buffer.
Nevertheless, $\hat{g}$ still retains valuable gradient information, as evidenced by the substantial performance gain when combining $\hat{g}$ with replay in UCIT.

Additionally, this ablation study reveals the impact of replay buffer size. For UCIT with significant distribution shifts, larger buffers yield considerable improvements, while for VQAv2 with homogeneous distributions, the benefits of increasing buffer size are limited.

\begin{figure}[t]
    \centering
    \begin{subfigure}{0.23\textwidth}  
        \centering
        \includegraphics[width=\textwidth]{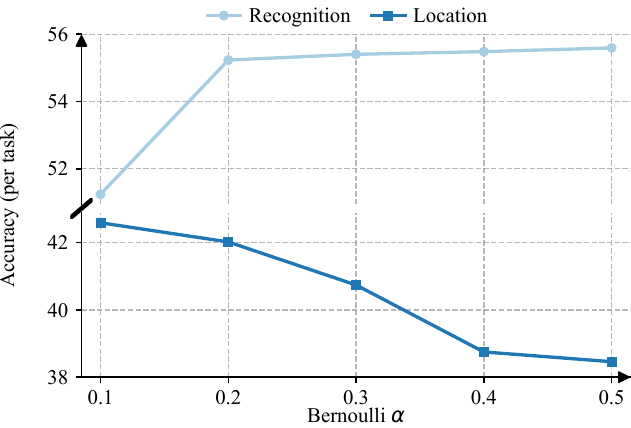}  
        \caption{VQAv2}
        \label{fig:ablation_alpha_vqav2_task2}
    \end{subfigure}
    \begin{subfigure}{0.23\textwidth}  
        \centering
        \includegraphics[width=\textwidth]{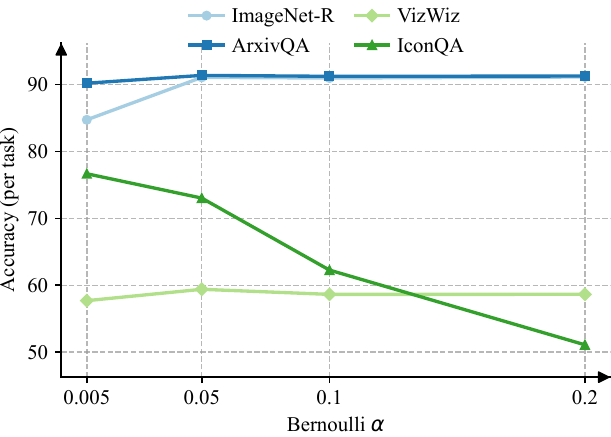}  
        \caption{UCIT}
        \label{fig:ablation_alpha_ucit_task4}
    \end{subfigure}
    
    \caption{Result of ablation on the impact of hyperparameter $\alpha$.}
    \label{fig:ablation_alpha}
\end{figure}

\noindent\textbf{Impact of Hyperparameter $\alpha$.} To validate the impact of hyperparameter $\alpha$ on model plasticity and stability, we conduct ablation experiments using different $\alpha$ values on specific tasks across both datasets. For VQAv2, where the replay buffer $\mathcal{M}$ has minimal influence, we adopt the $\hat{g}$-only approach (all following experiments on VQAv2 in this paper are performed without the replay buffer $\mathcal{M}$) and test $\alpha \in \{0.1, 0.2, 0.3, 0.4, 0.5\}$ during the learning of the second task (Location). For UCIT, we employ the full method and evaluate $\alpha \in \{0.005, 0.05, 0.1, 0.2\}$ during the fourth task (IconQA) learning phase. The results are depicted in Fig. \ref{fig:ablation_alpha}.
On VQAv2, we observe that as $\alpha$ increases—corresponding to more frequent gradient updates with $\hat{g}$—the accuracy of old tasks improves, while the accuracy of the new task gradually declines. This demonstrates $\alpha$'s role in balancing plasticity and stability. On UCIT, the effect is more pronounced: smaller $\alpha$ values clearly enhance plasticity for the new task. However, for old tasks, increasing $\alpha$ does not uniformly improve stability across all tasks, but excessively small $\alpha$ values consistently degrade the stability of old tasks. 

\begin{table}[t]
\centering
\caption{The FAA result of ablation on gradient scaling and Bernoulli sampling. The downward arrows indicate performance degradation compared to the full method.}
\label{tab:ablation_components}
\resizebox{\columnwidth}{!}{%
\begin{tabular}{cc|cc}
\toprule
\textbf{Gradient Scaling} & \textbf{Bernoulli Sampling} & \textbf{VQAv2} & \textbf{UCIT} \\
\midrule
\checkmark & \checkmark & 64.61 & 73.82 \\
$\times$ & \checkmark & 64.01\textsubscript{$\downarrow$0.60} & 65.24\textsubscript{$\downarrow$8.58} \\
\checkmark & $\times$ & 62.75\textsubscript{$\downarrow$1.86} & 59.02\textsubscript{$\downarrow$14.80} \\
\bottomrule
\end{tabular}%
}
\end{table}

\noindent\textbf{Ablation on Gradient Scaling and Bernoulli Sampling.} 
To further validate the importance of two key operations in our method—gradient scaling during gradient approximation and Bernoulli sampling for dynamic gradient updates—we conduct comprehensive ablation studies. Table \ref{tab:ablation_components} presents the performance comparison when these operations are selectively enabled or disabled. 
Specifically, disabling gradient scaling means directly using the raw directional vector between current and previous optimal parameters without scaling in Eq.~\ref{eq:g_hat_final} (i.e., $\theta - \theta^*_{1:t-1}$), while disabling Bernoulli sampling involves applying the approximated gradients $\hat{g}$ at every optimization step without stochastic sampling.
The results demonstrate that both components significantly impact the final performance. Regarding gradient scaling, its effect is more pronounced on UCIT with substantial distribution shifts, where performance decreases by 8.58\%, compared to only 0.6\% on VQAv2 with homogeneous distributions. For the Bernoulli sampling operation, it proves crucial across both datasets, with notably stronger impact than gradient scaling. Performance degrades by 14.8\% on UCIT and 1.86\% on VQAv2 when Bernoulli sampling is disabled. These findings indicate that both operations play more critical roles in scenarios with significant distribution shifts, while still providing measurable benefits even in more homogeneous scenarios.

\begin{figure}[t]
    \centering
    \begin{subfigure}{0.23\textwidth}  
        \centering
        \includegraphics[width=\textwidth]{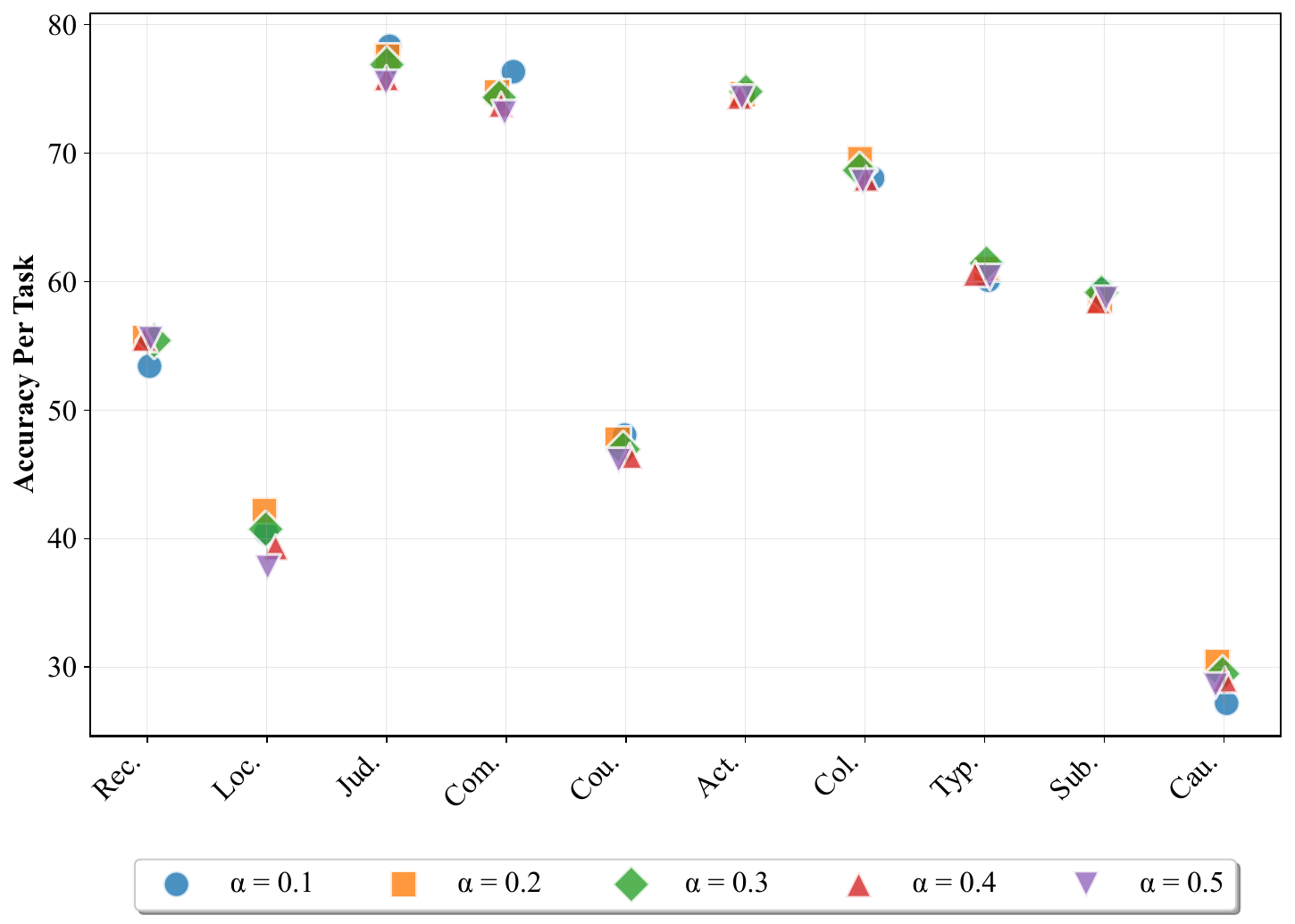}  
        \caption{VQAv2 (Accuracy per task)}
        \label{fig:sensetive_analysis_alpha_vqav2}
    \end{subfigure}
    \begin{subfigure}{0.23\textwidth}  
        \centering
        \includegraphics[width=\textwidth]{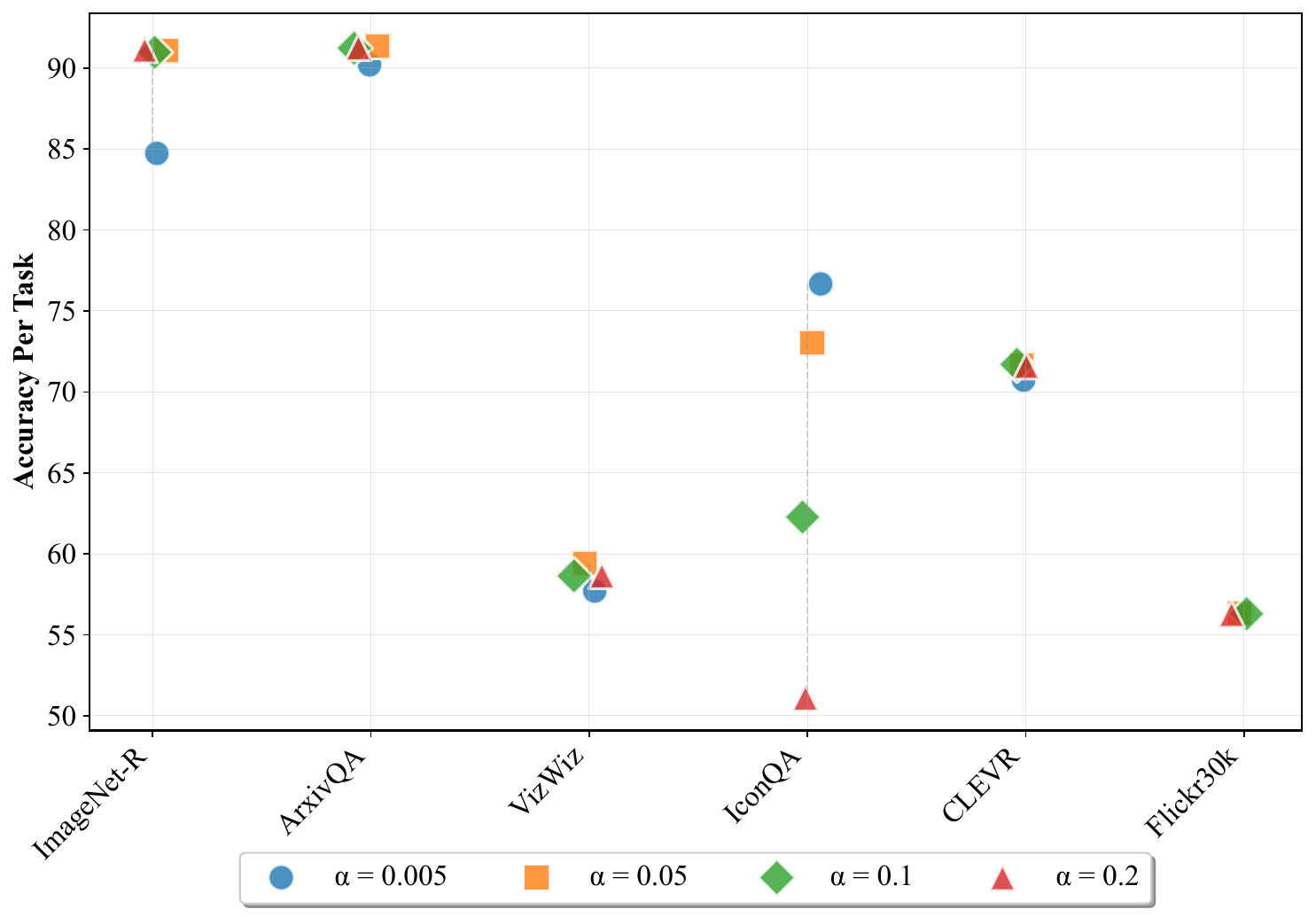}  
        \caption{UCIT (Accuracy per task)}
        \label{fig:sensetive_analysis_alpha_ucit}
    \end{subfigure}

    \vspace{0.5cm}

    \begin{subfigure}{0.23\textwidth}  
        \centering
        \includegraphics[width=\textwidth]{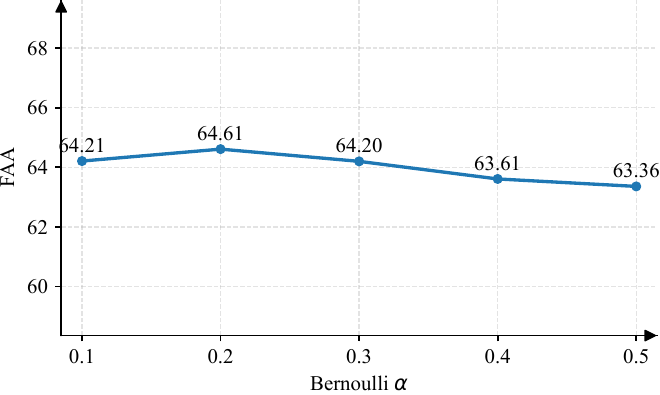}  
        \caption{VQAv2 (FAA)}
        \label{fig:sensetive_analysis_faa_vqav2}
    \end{subfigure}
    \begin{subfigure}{0.23\textwidth}  
        \centering
        \includegraphics[width=\textwidth]{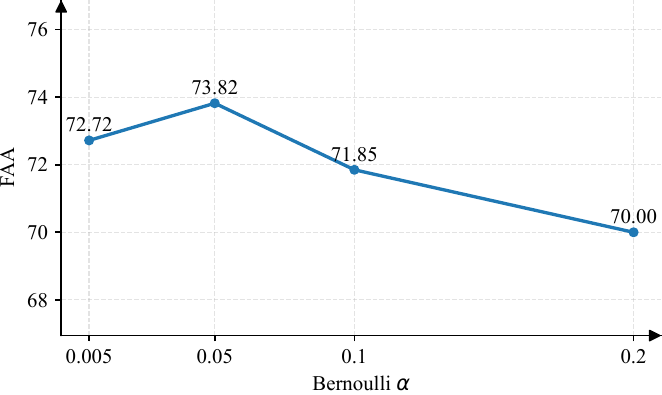}  
        \caption{UCIT (FAA)}
        \label{fig:sensetive_analysis_faa_ucit}
    \end{subfigure}
    
    \caption{Result of sensitivity analysis on the hyperparameter $\alpha$. Figure (a) and (b) show the per-task accuracy after completing all tasks on VQAv2 and UCIT datasets, respectively,  under different $\alpha$ parameter settings. (c) and (d) present the corresponding FAA on VQAv2 and UCIT datasets across varying $\alpha$ values.
 }
    \label{fig:sensetive_analysis}
\end{figure}

\subsection{Sensitivity Analysis of Hyperparameter $\alpha$}

During our ablation studies on $\alpha$, we observed significant performance fluctuations on the UCIT dataset when learning new tasks under different $\alpha$ values. To systematically investigate this phenomenon, we conduct a comprehensive sensitivity analysis examining how $\alpha$ affects the accuracy of each task and the FAA after completing continual instruction tuning all tasks.
For VQAv2, where all tasks share the same $\alpha$ configuration, we employ the $\hat{g}$-only approach and evaluate $\alpha \in \{0.1, 0.2, 0.3, 0.4, 0.5\}$ throughout the entire continual instruction tuning process. For UCIT, we maintain the full method setup and only vary $\alpha$ for the IconQA task within $\{0.005, 0.05, 0.1, 0.2\}$, while keeping $\alpha$ values for other tasks consistent with the main experiments.

The experimental results are summarized in Fig. \ref{fig:sensetive_analysis}. On VQAv2, which exhibits minimal distribution shifts, the performance remains relatively stable across different $\alpha$ values. As shown in Fig. \ref{fig:sensetive_analysis_alpha_vqav2}, varying $\alpha$ values cause only minor performance fluctuations across individual tasks, with the overall FAA varying by merely 1.25\% (Fig. \ref{fig:sensetive_analysis_faa_vqav2}). In contrast, UCIT with significant distribution variations demonstrates substantially higher sensitivity. When training the IconQA task with different $\alpha$ values, considerable performance fluctuations on IconQA are observed (Fig. \ref{fig:sensetive_analysis_alpha_ucit}), resulting in FAA variations of up to 3.81\% (Fig. \ref{fig:sensetive_analysis_faa_ucit}). This finding indirectly suggests that our gradient approximation $\hat{g}$ diverges from the true old task gradients when dealing with datasets featuring large distribution discrepancies, making the method more sensitive to the frequency control parameter $\alpha$.

Furthermore, both datasets exhibit a consistent trend: larger $\alpha$ values do not necessarily yield better performance. Excessively large $\alpha$ values significantly impair plasticity for new tasks, thereby degrading overall performance, while extremely small $\alpha$ values consistently damage stability for old tasks.
\section{Conclusion and Limitation}

In this paper, we introduce a novel insight into catastrophic forgetting by reformulating knowledge preservation as a gradient approximation problem. To approximate the gradient, we propose a dynamic gradient guidance method that utilizes optimal parameters from previous tasks as directional guidance. The approximated gradient can be further combined with real gradients from replay samples to form a more accurate estimation of old tasks' gradients. Additionally, we develop a Bernoulli sampling-based dynamic gradient update strategy to effectively control the stability-plasticity trade-off during continual instruction tuning.

Our method has been evaluated on two distinct MCIL datasets featuring similar and divergent data distributions, demonstrating its effectiveness and robustness. 
However, our experiments also reveal certain limitations: in scenarios with significant distribution shifts, the method exhibits higher dependency on replay buffers, necessitating additional storage requirements. Moreover, under such conditions, the approach shows increased sensitivity to the hyperparameter controlling gradient update frequency. Future work will focus on addressing these limitations through more adaptive gradient approximation techniques.

\section{Acknowledgments}
This work was supported by the National Key Research and Development Program of China (Grant No. 2024YFB3309400), the National Natural Science Foundation of China (Grant No. 62277011), the Open Research Fund from Guangdong Laboratory of Artificial Intelligence and Digital Economy (SZ) (Grant No.GML-KF-24-18), the Project of Chongqing MEITC (Grant No. YJX-2025001001009), the CAAI-CANN Open Fund and the CAS Project for Young Scientists in Basic Research (YSBR-083). This work was developed on OpenI Community.
{
    \small
    \bibliographystyle{ieeenat_fullname}
    
}

\clearpage
\setcounter{page}{1}
\setcounter{section}{0}
\setcounter{figure}{0}
\maketitlesupplementary

\begin{figure*}[t]
    \centering 
        \centering
        \includegraphics[width=\textwidth]{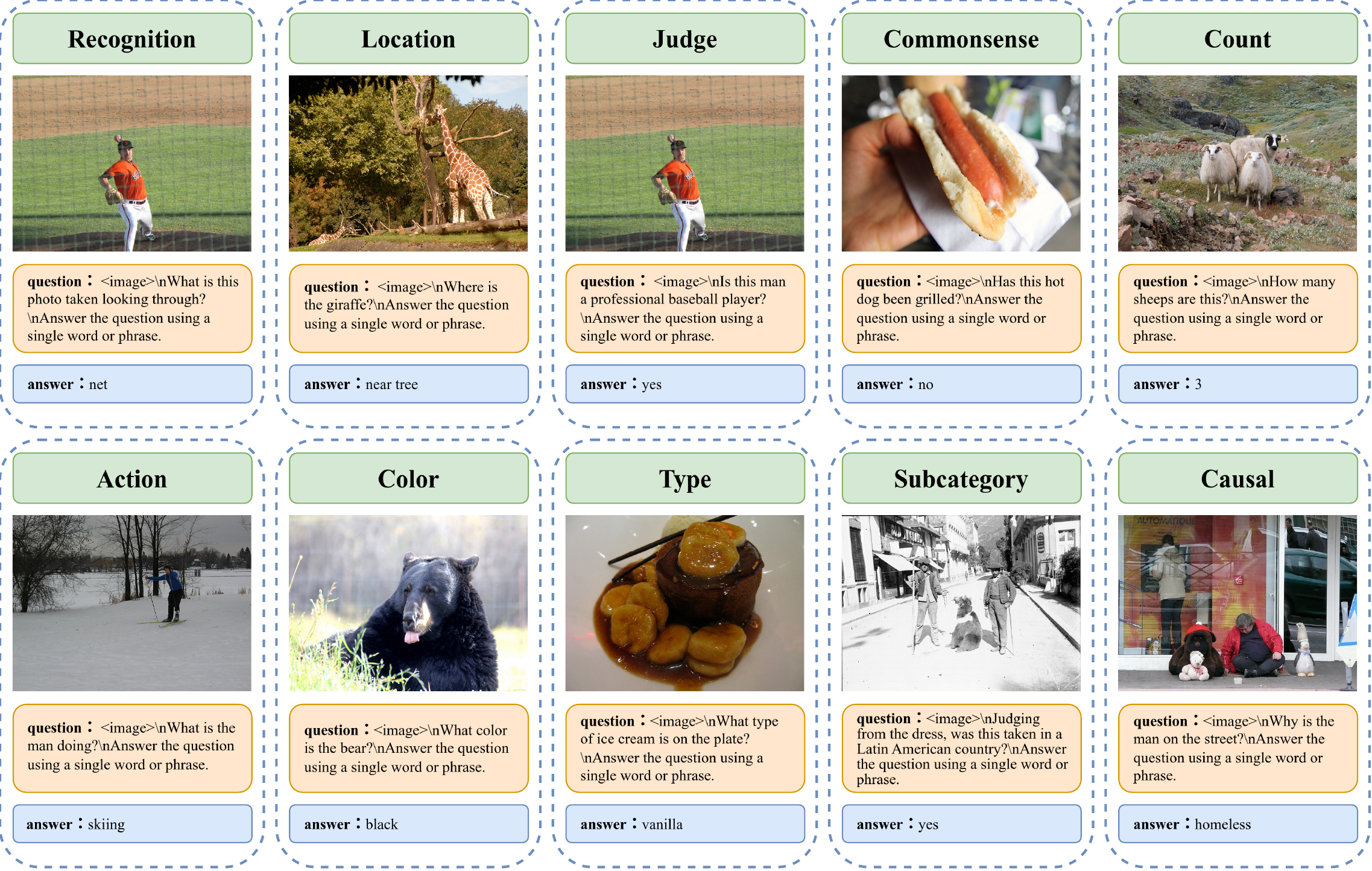}  

    \caption{Illustration of VQAv2 dataset.}
    \label{fig:vqav2_dataset}
\end{figure*}

\section{Dataset Distribution Analysis}

In the experimental section of this paper, we observed that our algorithm exhibits distinct characteristics on datasets with similar versus disparate image distributions. To provide an intuitive illustration of these distributional differences across datasets, we visualize the three datasets employed in our study—VQAv2, UCIT, and our custom-designed dataset—to visually demonstrate the variations in their image distributions.

\subsection{VQAv2 Dataset}
The VQAv2 dataset is constructed based on the MS-COCO dataset, which consists of real-world photographs capturing diverse everyday scenes and objects. These images exhibit rich textual information and natural visual characteristics, with all samples in each task residing in a similar distribution space (see Fig. \ref{fig:vqav2_dataset}). 
Furthermore, different tasks within VQAv2 often share identical image data across various question-answer pairs (see the Recognition and Judge task in Fig. \ref{fig:vqav2_dataset}), resulting in minimal distribution shifts between tasks. 

\subsection{UCIT Dataset}
The UCIT benchmark comprises six distinct sub-datasets with substantial distribution discrepancies:

\begin{itemize}
    \item \textbf{ImageNet-R}: Contains various artistic and synthetic renditions of ImageNet classes, including paintings, sketches, and sculptures, representing a significant domain shift from natural images.
    
    \item \textbf{ArxivQA}: Comprises scientific figures and diagrams extracted from academic papers, featuring schematic representations and specialized visualizations.
    
    \item \textbf{VizWiz}: Consists of images captured by blind individuals using mobile phones, often containing practical everyday objects with varying quality and unconventional perspectives.
    
    \item \textbf{IconQA}: Features iconographic images and symbolic representations, characterized by simplified graphics and abstract visual elements.
    
    \item \textbf{CLEVR}: Utilizes synthetically generated 3D scenes with geometric shapes, exhibiting clean backgrounds and programmed object arrangements.
    
    \item \textbf{Flickr30k}: Contains natural photographs from the Flickr platform, depicting real-world scenes with diverse contextual elements.
\end{itemize}

\begin{figure*}[t]
    \centering 
        \centering
        \includegraphics[width=\textwidth]{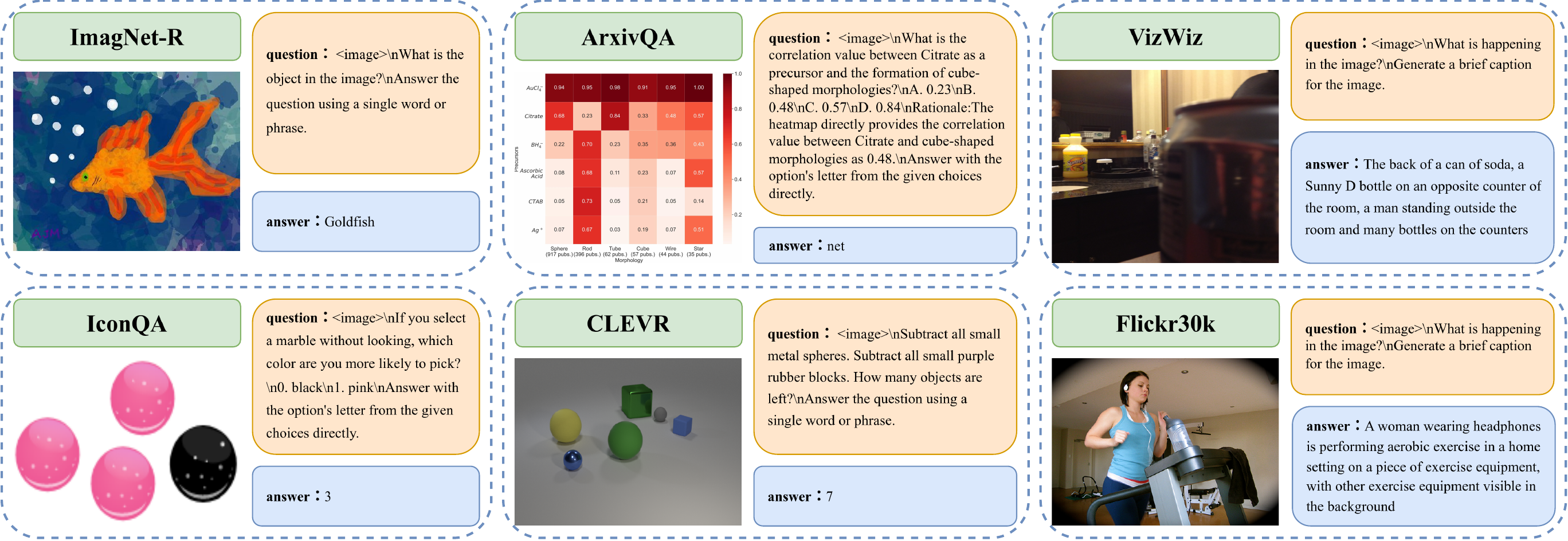}  

    \caption{Illustration of UCIT dataset.}
    \label{fig:ucit_dataset}
\end{figure*}

From Fig. \ref{fig:ucit_dataset}, we can observe substantial differences in image sources, visual characteristics, and content domains across these six sub-datasets, which result in significant distribution shifts and make UCIT a challenging benchmark.

\subsection{Custom Dataset}
\begin{figure*}[t]
    \centering 
        \centering
        \includegraphics[width=\textwidth]{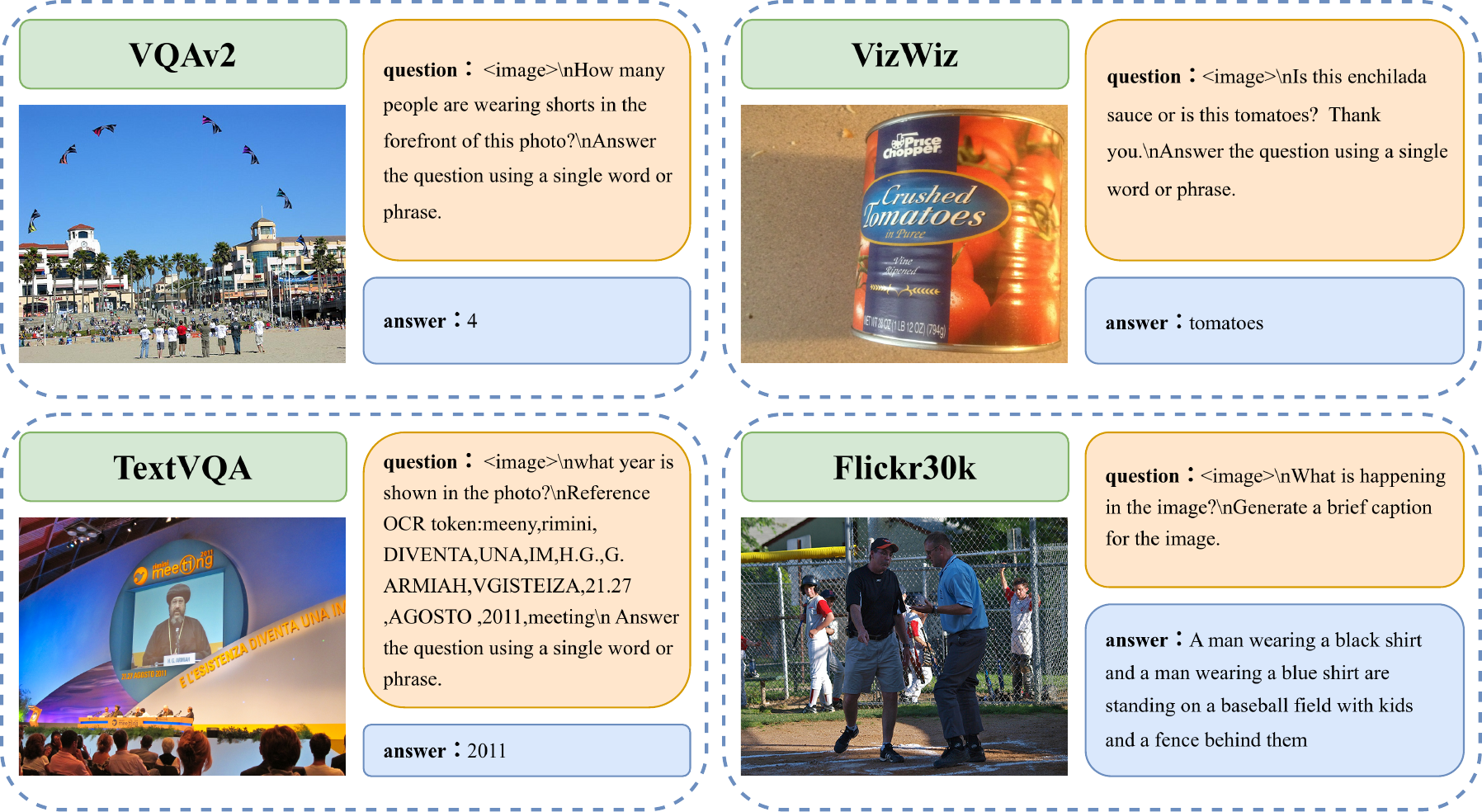}  

    \caption{Illustration of our custom dataset.}
    \label{fig:custom_dataset}
\end{figure*}

In the ablation study on gradient approximation, to verify that visual data distribution differences affect our method's dependency on replay data, we construct a custom dataset sequence (VQAv2 $\rightarrow$ VizWiz $\rightarrow$ TextVQA $\rightarrow$ Flickr30k). 
The TextVQA dataset focuses on visual question answering tasks that require reading and understanding text within images to answer questions about textual content in visual scenes. 
From Fig.~\ref{fig:custom_dataset}, it can be observed that although these four sub-datasets exhibit certain variations in specific visual properties, they primarily consist of real-world photographic data with rich textual information and natural scene representations. Compared to the UCIT benchmark, these datasets share more similar distribution characteristics due to their common origin in photographic imagery and comparable visual texture complexity.

\section{More Implementation Details}

\textbf{Model Architecture and Fine-tuning Strategy.}
Our approach is built upon the LLaVA (Large Language-and-Vision Assistant) model, which represents a pioneering framework for integrating visual and linguistic understanding. LLaVA connects a pre-trained vision encoder with a large language model through a carefully designed projection layer that aligns visual features with the language model's semantic space. This architecture enables the model to process multimodal inputs by first encoding visual information through the vision encoder, projecting these features into the language model's embedding space, and then jointly reasoning about visual and textual information using the language model's transformer blocks.

For parameter-efficient fine-tuning, we employ Low-Rank Adaptation (LoRA), a technique that approximates weight updates through low-rank decomposition. Specifically, for a pre-trained weight matrix $W_0 \in \mathbb{R}^{d \times k}$, LoRA constrains its update by representing it as the product of two low-rank matrices:
\begin{equation}
    W = W_0 + \Delta W = W_0 + BA
\end{equation}
where $B \in \mathbb{R}^{d \times r}$, $A \in \mathbb{R}^{r \times k}$, and the rank $r \ll \min(d,k)$. During training, only $A$ and $B$ are updated while $W_0$ remains frozen, significantly reducing the number of trainable parameters.


\textbf{Training Configuration.}
All experiments were conducted with a consistent batch size of 32 across both datasets and tasks. For the VQAv2 dataset, all subtasks except for the Causal task were trained for a single epoch, as this configuration provided sufficient convergence while minimizing computational overhead. The Causal task, which contains significantly fewer training samples compared to other subtasks, was trained for 4 epochs to ensure adequate learning. Similarly, all tasks in the UCIT dataset were trained for a single epoch to maintain consistency in training strategy across datasets. This differential training strategy ensures balanced optimization across all tasks regardless of their dataset sizes. The learning rate was maintained at $1\times10^{-4}$ throughout the training process, with linear warmup and cosine decay scheduling applied for stable convergence. Each experiment is conducted over three trials.

\end{document}